\documentclass[sigconf,nonacm]{acmart}
\usepackage{xcolor}
\usepackage{amsmath}
\usepackage{graphicx} 
\usepackage{booktabs}
\usepackage{tabularx}

\AtBeginDocument{%
  }
\setcopyright{acmlicensed}
\copyrightyear{2018}
\acmYear{2018}
\acmDOI{XXXXXXX.XXXXXXX}
\acmConference[Conference acronym 'XX]{Make sure to enter the correct
 conference title from your rights confirmation email}{June 03--05,
 2018}{Woodstock, NY}
\acmISBN{978-1-4503-XXXX-X/2018/06}

\newcommand{\benchmarkname}{MLB}

\begin{document}

\title{\benchmarkname{}: A Scenario-Driven Benchmark for Evaluating Large Language Models in Clinical Applications}




\author{%
  Qing He\textsuperscript{1,*} 
  Dongsheng Bi\textsuperscript{1,*}, 
  Jianrong Lu\textsuperscript{2,1,*}, 
  Minghui Yang\textsuperscript{1}, 
  Zixiao Chen\textsuperscript{1}, 
  Jiacheng Lu\textsuperscript{1}, 
  Jing Chen\textsuperscript{1}, 
  Nannan Du\textsuperscript{1}, 
  Xiao Cu\textsuperscript{1}, 
  Sijing Wu\textsuperscript{3}, 
  Peng Xiang\textsuperscript{4}, 
  Yinyin Hu\textsuperscript{3}, 
  Yi Guo\textsuperscript{3}, 
  Chunpu Li\textsuperscript{3}, 
  Shaoyang Li\textsuperscript{1}, 
  Zhuo Dong\textsuperscript{1}, 
  Ming Jiang\textsuperscript{1}, 
  Shuai Guo\textsuperscript{1}, 
  Liyun Feng\textsuperscript{1}, 
  Jin Peng\textsuperscript{1}, 
  Jian Wang\textsuperscript{1}, 
  Jinjie Gu\textsuperscript{1}, 
  Junwei Liu\textsuperscript{1,5, \textdagger}
}


\affiliation{%
  \institution{%
    \textsuperscript{1} Ant Group, Hangzhou, China\\
    \textsuperscript{2} Zhejiang University, Hangzhou, China\\
    \textsuperscript{3} Health Information Center of Zhejiang Province, Hangzhou, China\\
    \textsuperscript{4} Department of AI and IT, The Second Affiliated Hospital, School of Medicine, Zhejiang University, Hangzhou, China\\
    \textsuperscript{5} School of Software and Microelectronics, Peking University, Beijing, China \\
    * The first three authors contributed equally to this work; \textdagger Corresponding Author.
    \country{}
  }
}

\email{{bidongsheng.bds, minghui.ymh, bobblair.wj}@antgroup.com, jrong.alvin@gmail.com, liujunwei@stu.pku.edu.cn}

\begin{abstract}
The proliferation of Large Language Models (LLMs) presents transformative potential for healthcare, yet practical deployment is hindered by the absence of frameworks that assess real-world clinical utility. Existing benchmarks test static knowledge, failing to capture the dynamic, application-oriented capabilities required in clinical practice. To bridge this gap, we introduce a \textbf{M}edical\textbf{ L}LM \textbf{B}enchmark  \textbf{\benchmarkname{}}, a comprehensive benchmark evaluating LLMs on both foundational knowledge and scenario-based reasoning. \benchmarkname{} is structured around five core dimensions: Medical Knowledge (MedKQA), Safety and Ethics (MedSE), Medical Record Understanding (MedRU), Smart Services (SmartServ), and Smart Healthcare (SmartCare). The benchmark integrates 22 datasets (17 newly curated) from diverse Chinese clinical sources, covering 64 clinical specialties. Its design features a rigorous curation pipeline involving \textbf{300 licensed physicians}. Besides, we provide a scalable evaluation methodology, centered on a specialized judge model trained via Supervised Fine-Tuning (SFT) on expert annotations. Our comprehensive evaluation of 10 leading models reveals a critical translational gap: while the top-ranked model, Kimi-K2-Instruct (77.3\% accuracy overall), excels in structured tasks like information extraction (87.8\% accuracy in MedRU), performance plummets in patient-facing scenarios (61.3\% in SmartServ). Moreover, the exceptional safety score (90.6\% in MedSE) of the much smaller Baichuan-M2-32B highlights that targeted training is equally critical. Our specialized judge model, trained via SFT on a 19k expert-annotated medical dataset, achieves 92.1\% accuracy, an F1-score of 94.37\%, and a Cohen’s Kappa of 81.3\% for human–AI consistency, validating a reproducible and expert-aligned evaluation protocol. \benchmarkname{} thus provides a rigorous framework to guide the development of clinically viable LLMs.
\end{abstract}

\ccsdesc[500]{Computing methodologies~Natural language generation}
\ccsdesc[500]{Applied computing~Health care information systems}
\keywords{Large Language Models, Benchmark, Medical AI, Healthcare, Evaluation, Supervised Fine-Tuning}
\received{23 November 2025}
\maketitle
\renewcommand{\shortauthors}{Qing He et al.}

\section{Introduction}
The rapid advancement of LLMs holds immense promise for revolutionizing healthcare, with potential applications ranging from clinical decision support \cite{singhal2023large, thirunavukarasu2023large} to specialized tasks like mental health analysis \cite{zhang2024mentallama,lu2024depriving,lu2025parasolver} and patient message triage \cite{gove2024triage}. However, translating this potential into safe and effective clinical practice requires overcoming significant challenges related to professional accuracy, adaptability to complex scenarios, and adherence to stringent safety and ethical standards \cite{lee2023benefits, fan2024mmadapt, cima2025contextualized}. For example, current benchmarks for medical LLMs, such as those based on standardized licensing exams \cite{jin2019pubmedqa, pal2022medmcqa}, primarily assess foundational medical knowledge through multiple-choice questions. While valuable, these evaluations fail to measure the models' practical utility in authentic clinical contexts \cite{zhang2024howmuch} and, critically, lack a scalable and reliable method for assessing the open-ended, nuanced reasoning required in clinical interactions. This highlights an urgent need for a comprehensive, application-oriented benchmark to guide the development and deployment of LLMs in healthcare.

To address this critical gap, we developed the \benchmarkname{}, a benchmark designed to systematically evaluate the end-to-end capabilities of LLMs in healthcare within the Chinese clinical healthcare context. The \benchmarkname{} moves beyond static knowledge assessment to prioritize performance in diverse, realistic clinical scenarios, thereby assessing the breadth, depth, and practical relevance of current models. Our objective is to provide a scientific and objective framework for quantifying the reliability, applicability, and safety of LLMs, fostering the development of high-quality, trustworthy medical AI, aligning with the broader goal of leveraging web and AI technologies for social good.

The main contributions of this work are as follows:
\begin{enumerate}
    \item We introduce \benchmarkname{} a novel, scenario-driven benchmark constructed from challenging, real-world data sources, including multi-turn physician-patient dialogues from web-based telehealth platforms and authentic clinical records. Its expert-driven (300 physicians) curation and difficulty grading provide a robust assessment of practical clinical utility.
    \item We propose and validate a scalable evaluation methodology for complex, open-ended clinical tasks. By applying SFT to a large corpus of expert-annotated judgments, we developed a specialized ``judge" model that achieves high-fidelity, reproducible, and cost-effective evaluation at scale.
    \item We conducted a comprehensive evaluation of 10 leading LLMs. This analysis reveals a critical translational gap: top models like Claude-4-Sonnet achieve high proficiency in structured tasks (88.8\% in MedRU), yet the best-performing model, Kimi-K2-Instruct, only scores 61.3\% in patient-facing scenarios (SmartServ). Our SFT-judge model proves essential for this analysis, achieving 92.1\% accuracy and 81.3\% Cohen's Kappa, demonstrating near-human agreement.
\end{enumerate}

\vspace{-4mm}
\section{Related Work} The evaluation of LLMs~\cite{zhou2024llm,lu2024depriving} in the medical domain has evolved from foundational knowledge tests to more comprehensive assessments. Early and prominent benchmarks were largely adapted from existing medical examinations. For instance, MedExQA \cite{kim2024medexqa} utilized questions from the United States Medical Licensing Examination (USMLE), establishing a standard for assessing clinical knowledge. Similarly, PubMedQA \cite{jin2019pubmedqa} focused on biomedical research questions requiring yes/no/maybe answers, while MedMCQA \cite{pal2022medmcqa} used multiple-choice questions from Indian medical entrance exams. While these benchmarks are crucial for gauging a model's knowledge base, their format does not capture the complexities of clinical interaction and reasoning.

Recognizing these limitations, subsequent efforts aimed for broader evaluations. The MultiMedQA benchmark \cite{singhal2023large} integrated several existing datasets and introduced HealthSearchQA, a dataset of commonly searched consumer health questions, to assess the model's ability to provide layperson-friendly information. Another notable effort, MedBench \cite{li2024medbench}, expanded the scope by including tasks related to clinical notes, diagnostics, and patient conversations. These benchmarks represent a significant step towards more holistic evaluation.

A recent, concurrent effort is OpenAI's HealthBench \cite{arora2025healthbenchevaluatinglargelanguage}, which also emphasizes real-world scenarios through 5,000 simulated dialogues. HealthBench employs a distinct methodology, utilizing a generalist model (GPT-4.1) to score responses against unique, physician-authored rubrics for each dialogue. While \benchmarkname{} shares the goal of scenario-based evaluation, it provides a more diversified assessment framework, integrating 22 datasets across five distinct dimensions (MedKQA, MedSE, MedRU, SmartServ, SmartCare), rather than focusing solely on dialogue quality. Methodologically, our SFT-trained judge model is developed as a scalable proxy for human adjudication on disputed, open-ended tasks, a different approach from HealthBench's rubric-based model evaluation. Furthermore, \benchmarkname{}'s scenarios are derived from authentic clinical records and web-based telehealth data, complementing HealthBench's synthetically generated interactions.

However, a critical gap remains in the systematic evaluation of LLMs in applied, scenario-based contexts that simulate real-world clinical workflows. While MedBench \cite{li2024medbench} incorporates clinical dialogue tasks, its scenarios often lack the complexity of real-world data streams. \benchmarkname{}'s \textit{SceneCap} dimension provides a more substantial differentiation by simulating the entire clinical workflow. Our tasks are not isolated conversations but are constructed from authentic, multi-turn patient-provider interactions sourced from web-based telehealth platforms and anonymized electronic health records (EHRs).

This contrasts with benchmarks like MultiMedQA \cite{singhal2023large}, which merely aggregated existing datasets, or earlier web-tools focused on information retrieval over clinical reasoning \cite{simmons2010clinical}. \benchmarkname{}'s innovation lies in its expert-driven, synthesis-oriented data curation. Our clinical experts (including web-health practitioners) fused disparate, cross-source medical records (e.g., lab reports, notes) to create novel, complex test cases. This methodology explicitly assesses an LLM's ability to handle the fragmented, conflicting data endemic to clinical practice—a dimension previous benchmarks failed to measure. Our use of web-telehealth data aligns with recent work scaling medical AI via web technologies \cite{guo2022webformer, hsu2009development, giannakos2010evaluation}. While parallel efforts address other modalities \cite{fan2025towards}, robust conversational AI evaluation remains a key challenge.

Critically, while these benchmarks introduce more complex tasks, they often rely on simplified metrics (e.g., string matching) or costly, non-scalable human evaluation for open-ended questions. A significant methodological gap remains in the scalable and reproducible scoring of nuanced clinical reasoning, a gap we address with our SFT-based evaluation protocol.

\section{The \benchmarkname{} Benchmark}
\begin{figure*}[h]
 \centering
 \includegraphics[width=\linewidth]{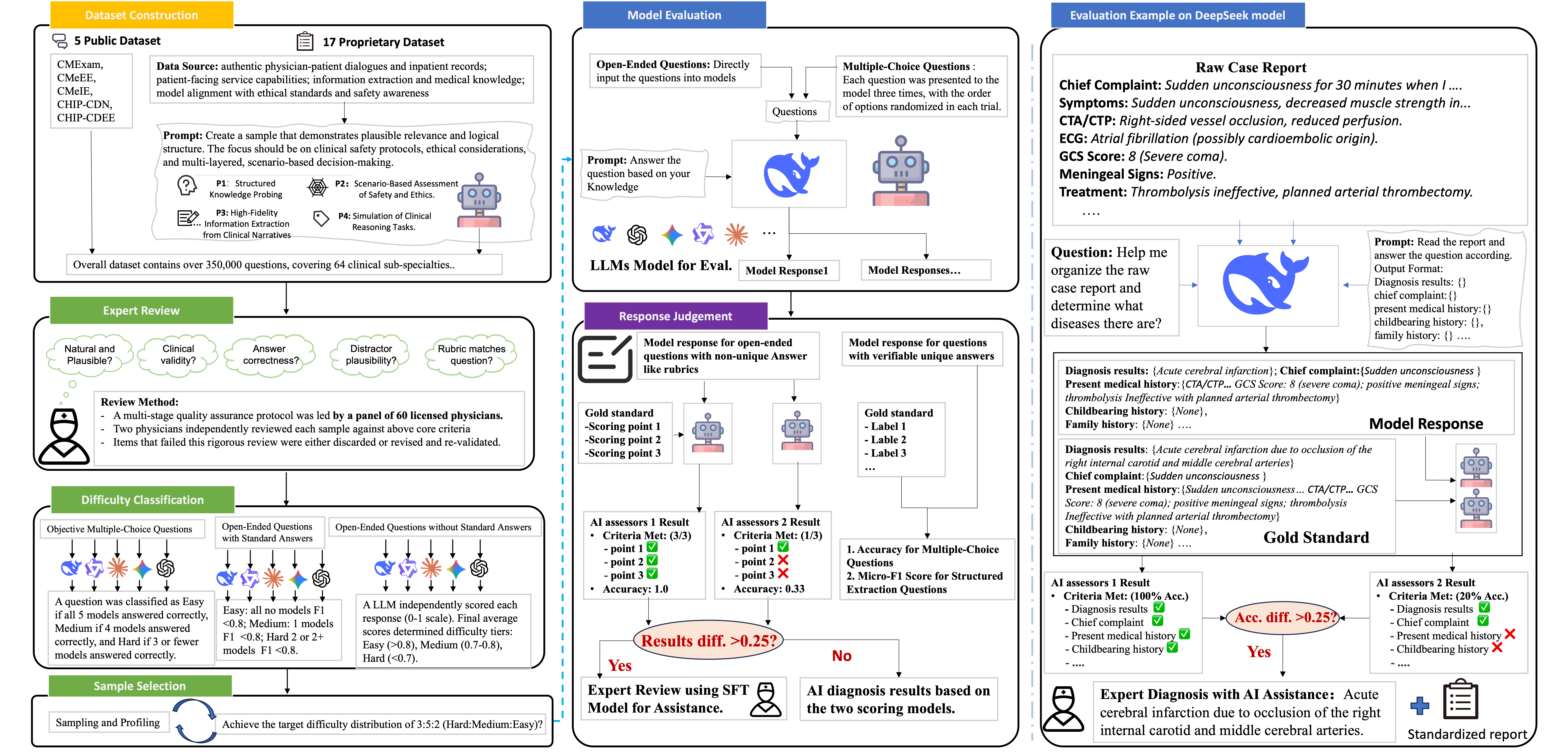}
 \caption{Pipeline of Dataset Curation and Evaluation. The diagram outlines the end-to-end process for creating and evaluating the \benchmarkname{}. It begins with data acquisition from diverse sources (e.g., EHRs, web-based dialogues), proceeds through a multi-paradigm generation pipeline (P1-P4), and undergoes rigorous expert review by 300 licensed physicians. The evaluation phase uses a hybrid scoring protocol in response judgement, culminating in our SFT-based judge model for resolving ambiguities.}
 \label{fig:curation_pipeline}
\end{figure*}

The \benchmarkname{} is designed as a multi-layered evaluation system that assesses LLMs from foundational knowledge to practical application. A key challenge in this design is the reliable assessment of the ``SceneCap" domain, which relies on open-ended clinical scenarios. We address this through a novel evaluation methodology (detailed in Section~\ref{sec:eval_methodology}) centered on a specialized SFT-trained judge model. The benchmark's hierarchical structure, detailed dataset composition, rigorous curation process, and sophisticated evaluation methodology collectively ensure a comprehensive and reliable assessment. Figure~\ref{fig:curation_pipeline} shows the curation and evaluation pipeline.

\begin{figure}[h]
 \centering
 \includegraphics[width=\linewidth]{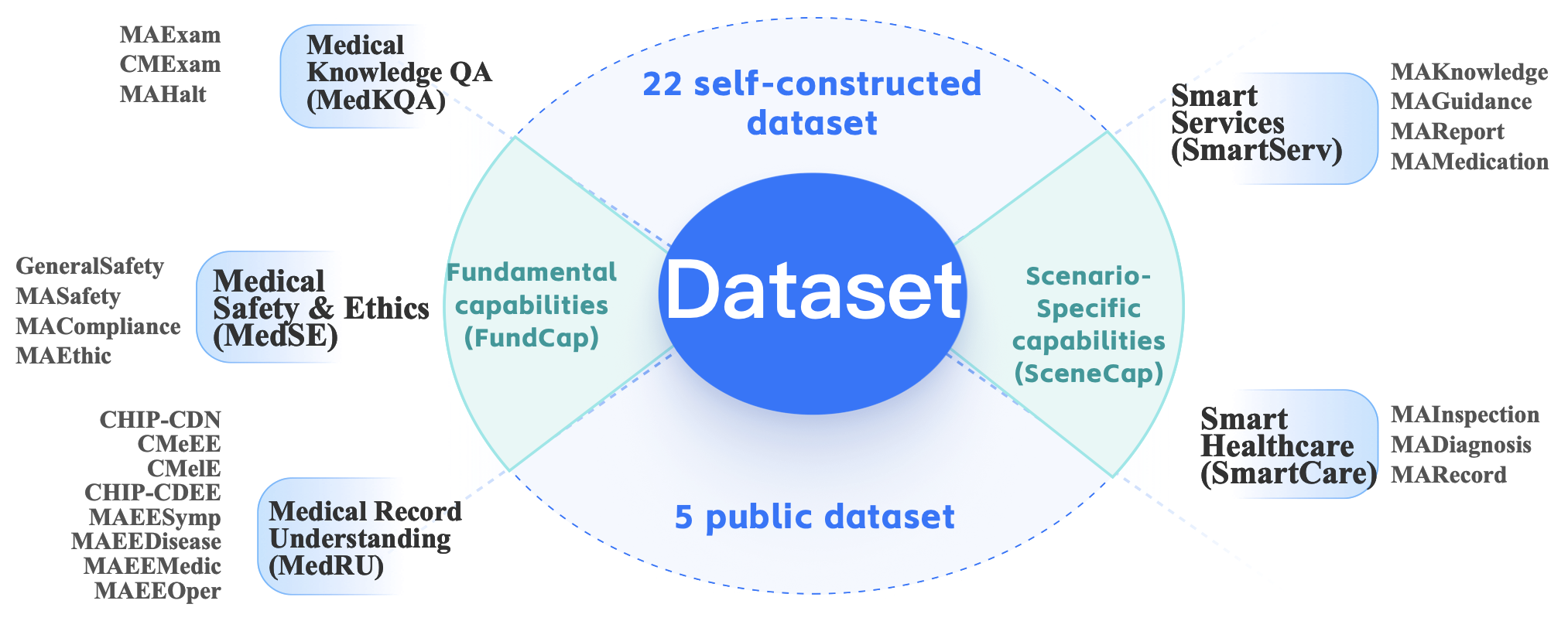}
 \caption{Overview of the \benchmarkname{} Benchmark. This figure illustrates the benchmark's hierarchical structure, divided into Fundamental Capabilities (FundCap) and Scenario-based Capabilities (SceneCap), which are further broken down into five core dimensions (MedKQA, MedSE, MedRU, SmartServ, SmartCare).}
 \label{fig:overall_introduction}
\end{figure}

\vspace{-8mm}
\subsection{Evaluation Dimensions}
The benchmark framework is organized into a two-level hierarchy. At the highest level, we distinguish between \textbf{Fundamental Capabilities (FundCap)} and \textbf{Scenario-based Capabilities (SceneCap)}. These are further broken down into five core dimensions, as illustrated in Figure~\ref{fig:overall_introduction}.
\begin{itemize}
 \item \textbf{Fundamental Capabilities (FundCap)}: This domain assesses the core knowledge and processing abilities of an LLM in the medical field. It is divided into three dimensions:
 \begin{enumerate}
 \item \textbf{Medical Knowledge Question Answering (MedKQA)}: Evaluates the accuracy and breadth of an LLM's medical knowledge and its ability to mitigate hallucinations.
 \item \textbf{Medical Safety \& Ethics (MedSE)}: Measures the model's adherence to safety protocols, regulatory compliance, and ethical principles in a medical context.
 \item \textbf{Medical Record Understanding (MedRU)}: Assesses the model's capacity to process and extract structured information from unstructured clinical text, such as electronic health records (EHRs).
 \end{enumerate}
 \item \textbf{Scenario-based Capabilities (SceneCap)}: This domain evaluates the model's performance in simulated real-world clinical applications. It is divided into two dimensions:
 \begin{enumerate}
 \item \textbf{Smart  Services (SmartServ)}: Focuses on patient-facing tasks, such as providing \textbf{Health Education}, \textbf{Department Guidance}, report interpretation, and medication instructions.
 \item \textbf{Smart Healthcare (SmartCare)}: Focuses on physician-support tasks, such as \textbf{Medical Inspection Recommendation}, assisting with \textbf{Diagnostic Assistance}, and \textbf{Medical Record Writing Assistance}.
 \end{enumerate}
\end{itemize}

\vspace{-5mm}
\subsection{Dataset Composition}
\begin{figure*}[h]
 \centering
 \includegraphics[width=\linewidth]{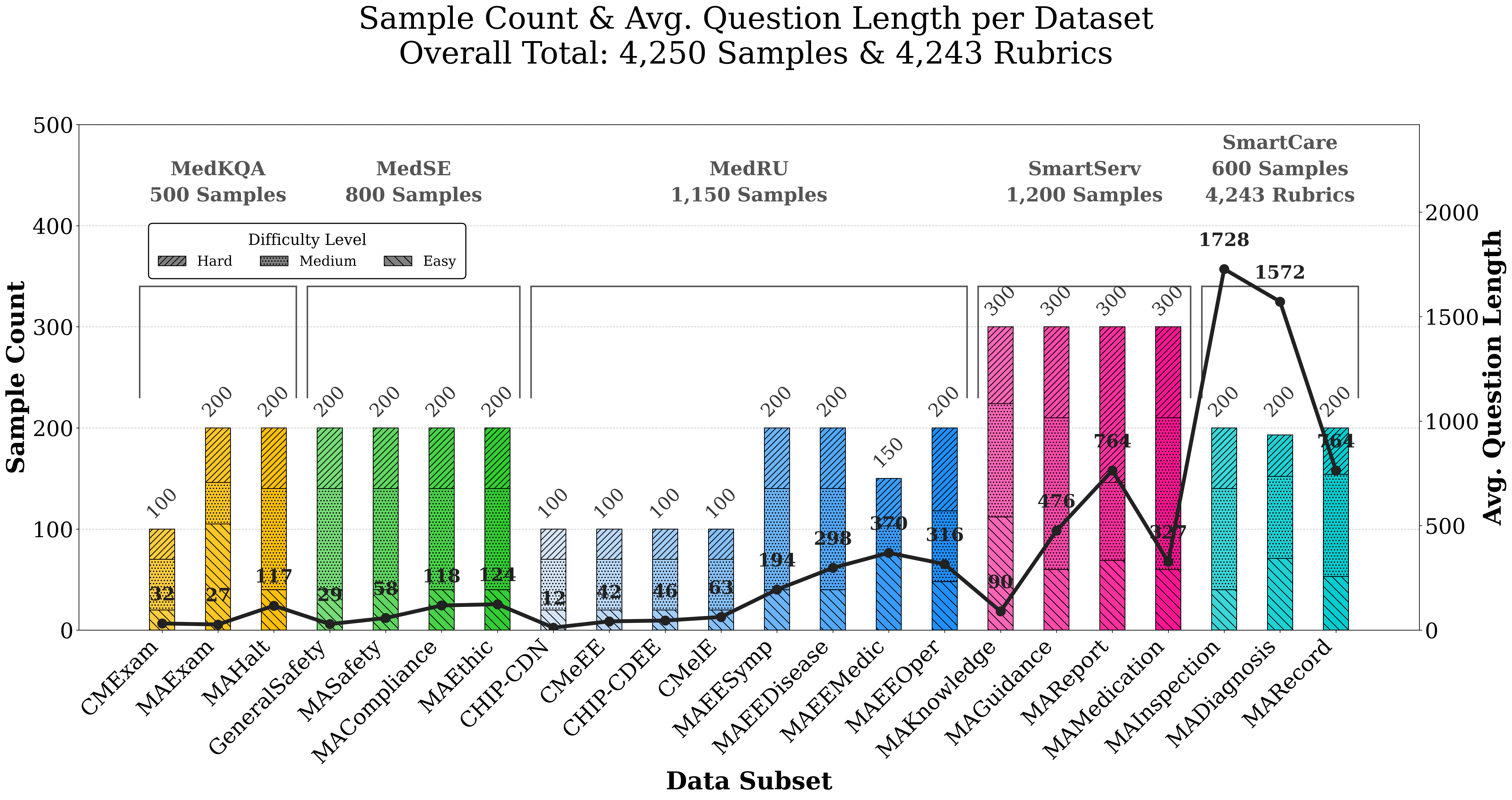}
 \caption{Overview of the dataset composition, including total sample size, distribution across dimensions, average question length, and difficulty level. The chart displays the distribution of \benchmarkname{}'s 22 datasets across the five core dimensions. It highlights the total sample size (4250 samples), the source of data (17 proprietary, 5 public), and the average token length of questions, indicating the complexity of the tasks.}
 \label{fig:dataset_composition}
\end{figure*}

To operationalize this framework, the \benchmarkname{} comprises 22 distinct datasets, including 17 newly constructed (proprietary) datasets and 5 established public datasets. These datasets are derived from a diverse range of sources, including clinical guidelines, medical textbooks, authentic EHRs, and physician-patient dialogues, ensuring robust and realistic evaluation scenarios. Figure~\ref{fig:dataset_composition} shows that the overall dataset contains over 350,000 questions, covering 64 clinical sub-specialties. Table~\ref{tab:dataset_overview} details the construction of the  dataset The specific specialty distribution is shown in Table~\ref{tab:department_distribution}. 

\begin{table*}[t]
 \caption{Overview of Datasets, Evaluation Methods and Sampling Strategy in the \benchmarkname{}. This comprehensive table details all 22 datasets within \benchmarkname{}, organized by their primary, secondary, and tertiary dimensions. It specifies the data source (Public/Proprietary), question type, scoring metric, original and final sampled item counts (N), and the key attributes (e.g., difficulty level, department) used for our stratified and proportional sampling strategy.}
 \label{tab:dataset_overview}
 \centering
 \resizebox{\textwidth}{!}{
 \begin{tabular}{lllllllccc}
 \toprule
 \textbf{Primary Dimension} & \textbf{Secondary Dimension} & \textbf{Tertiary Dimension Intro.} & \textbf{Dataset} & \textbf{Source} & \textbf{Type} & \textbf{Scoring Method} & \textbf{Original N} & \textbf{Sampled N} & \textbf{Sampling Key(s)} \\
 \midrule
 FundCap & MedKQA & Med. Knowledge & CMExam & Public & Multiple-Choice & Accuracy & 6811 & 100 & difficulty level \\
 &MedKQA &Med. Knowledge & MAExam & Proprietary & Multiple-Choice & Accuracy & 1000 & 200 & difficulty level \\
 &MedKQA & Med. Hallucination & MAHalt & Proprietary & Multiple-Choice & Accuracy & 5000 & 200 & type, difficulty level \\
 \cmidrule(l){2-10} 
 & MedSE & General Safety & GeneralSafety & Proprietary & Open-Ended & AI-Assisted w/ Expert Adj. & 3600 & 200 & difficulty level \\
 &MedSE & Medical Safety & MASafety & Proprietary & Multiple-Choice & Accuracy & 5000 & 200 & source, difficulty level \\
 &MedSE & Med. Compliance & MACompliance & Proprietary & Open-Ended & AI-Assisted w/ Expert Adj. & 3000 & 200 & risk type, difficulty level \\
 &MedSE & Medical Ethics & MAEthic & Proprietary & Open-Ended & AI-Assisted w/ Expert Adj. & 500 & 200 & risk type, difficulty level \\
 \cmidrule(l){2-10} 
 & MedRU & Term Normalization & CHIP-CDN & Public & Open-Ended & Micro-F1 & 2000 & 100 & department, difficulty level \\
 &MedRU & Info. Extraction & CMeEE & Public & Open-Ended & Micro-F1 & 5000 & 100 & department, difficulty level \\
 &MedRU & Info. Extraction & CHIP-CDEE & Public & Open-Ended & Micro-F1 & 384 & 100 & department, difficulty level \\
 &MedRU & Info. Extraction & CMeIE & Public & Open-Ended & Micro-F1 & 3585 & 100 & department, difficulty level \\
 &MedRU & Info. Extraction & MAEESymp & Proprietary & Open-Ended & Micro-F1 & 1040 & 200 & department, difficulty level \\
 &MedRU &Info. Extraction & MAEEDisease & Proprietary & Open-Ended & Micro-F1 & 716 & 200 & department, difficulty level \\
 &MedRU &Info. Extraction & MAEEMedic & Proprietary & Open-Ended & Micro-F1 & 393 & 150 & department, difficulty level \\
 &MedRU &Info. Extraction & MAEEOper & Proprietary & Open-Ended & Micro-F1 & 580 & 200 & department, difficulty level \\
 \midrule 
 SceneCap & SmartServ & Health Education & MAKnowledge & Proprietary & Multiple-Choice & Accuracy & 6983 & 300 & department, difficulty level \\
 &SmartServ & Department Guidance & MAGuidance & Proprietary & Multiple-Choice & Accuracy & 4837 & 300 & department, difficulty level \\
 &SmartServ & Report Interp. & MAReport & Proprietary & Multiple-Choice & Accuracy & 3000 & 300 & department, difficulty level \\
 &SmartServ & Medication Instruct. & MAMedication & Proprietary & Multiple-Choice & Accuracy & 5932 & 300 & department, difficulty level \\
 \cmidrule(l){2-10} 
 & SmartCare & Medical Inspection Recommendation & MAInspection & Proprietary & Open-Ended & AI-Assisted w/ Expert Adj. & 3000 & 200 & department, difficulty level \\
 &SmartCare & Diagnostic Assistance & MADiagnosis & Proprietary & Open-Ended & AI-Assisted w/ Expert Adj. & 3000 & 200 & department, difficulty level \\
 &SmartCare & Med. Record Writing Assistance & MARecord & Proprietary & Open-Ended & AI-Assisted w/ Expert Adj. & 3000 & 200 & department, difficulty level \\
 \bottomrule
 \end{tabular}
 }
\end{table*}

\begin{table}[htbp]
\centering
\caption{Distribution of data samples across 9 major clinical departments and 64 sub-specialties. This table provides a detailed breakdown of the samples annotated by clinical specialty. Internal Medicine (722 samples) is the most highly represented category.}
\label{tab:department_distribution}
\resizebox{\columnwidth}{!}{%
\begin{tabular}{ll}
\toprule
\textbf{Major Department (Samples)} & \textbf{Sub-specialties} \\
\midrule

Internal Medicine (722) & 
\begin{tabular}[t]{@{}l l l l l@{}}
Respiratory Medicine & Gastroenterology & Neurology & Cardiology & Hematology \\
Nephrology & Endocrinology & Rheumatology & Allergy and Immunology & Geriatrics Dept.
\end{tabular} \\
\midrule

Surgery (248) &
\begin{tabular}[t]{@{}l l l l@{}}
General Surgery & Neurosurgery & Orthopedic Surgery & Urology Surgery \\
Thoracic Surgery & Cardiovascular Surgery & Burn Surgery & Plastic Surgery
\end{tabular} \\
\midrule

Pediatrics (43) &
\begin{tabular}[t]{@{}l l l@{}}
Neonatology & Pediatric Cardiology & Pediatric Hematology \\
Pediatric Infectious Diseases Dept. & Pediatric Nephrology & Pediatric Genetics \\
Pediatric Gastroenterology & Pediatric Neurology & Pediatric Immunology \\
Pediatric Pulmonology & Pediatric Endocrinology &
\end{tabular} \\
\midrule

Obstetrics and Gynecology (119) & Gynecology \quad Women's Health \quad Obstetrics \quad Reproductive Endocrinology and Infertility \\
\midrule

Pediatric Surgery (24) &
\begin{tabular}[t]{@{}l l@{}}
Pediatric General Surgery & Pediatric Cardiothoracic Surgery \\
Pediatric Orthopedics & Pediatric Neurosurgery \\
Pediatric Urology & 
\end{tabular} \\
\midrule

Other Departments (366) &
\begin{tabular}[t]{@{}l l l@{}}
Ophthalmology & Infectious Diseases Dept. & Occupational Medicine \\
Otolaryngology & Tuberculosis (TB) Dept. & Anesthesiology \\
Stomatology & Endemic Diseases & Pain Management \\
Dermatology & Oncology & Critical Care Medicine \\
Sexually Transmitted Diseases Dept. & Emergency Medicine & Preventive Medicine \\
Medical Aesthetics Dept. & Rehabilitation Medicine & Nursing Dept. \\
Psychiatry Dept. & Sports Medicine & General Health
\end{tabular} \\
\midrule

Medical Technology Dept. (11) &
Pathology \quad Clinical Laboratory \quad Medical Photography \\

General Practice Dept. (86) & General Practice Dept. \\

Child Healthcare Dept. (8) & Child Healthcare Dept. \\

\bottomrule
\end{tabular}
} 
\end{table}

\subsection{Dataset Curation Process}
The integrity and quality of the benchmark are underpinned by a rigorous, human-in-the-loop curation process. This process combines the scalability of automated generation with the precision of expert verification.

\subsubsection{Dataset Preparation}
Our data curation strategy combined the use of established public benchmarks with the development of novel, purpose-built datasets to ensure both broad coverage and targeted assessment of advanced capabilities. This dual approach yielded a comprehensive collection of 22 datasets, comprising 5 public and 17 proprietary datasets.

\paragraph{Public Datasets}
To ground our benchmark in established tasks, we incorporated five widely recognized Chinese medical language processing datasets. These included: \textbf{CMExam} \cite{CMExam}, a dataset derived from the Chinese National Medical Licensing Examination; \textbf{CHIP-CDN} \cite{CHIP-CDN}, a clinical diagnosis normalization task; and a suite of information extraction datasets including \textbf{CMeEE} \cite{CMeEE} for named entity recognition, \textbf{CHIP-CDEE} \cite{CHIP-CDEE} for structured clinical event extraction, and \textbf{CMeIE} \cite{CMeIE} for identifying semantic relationships. These datasets provide a robust baseline for assessing foundational natural language understanding capabilities in the medical domain.

\paragraph{Proprietary Dataset Construction}
The 17 proprietary datasets were meticulously constructed to address gaps in existing benchmarks, particularly concerning clinical safety, ethics, and complex, scenario-based reasoning. Our construction methodology was centered on a sophisticated, prompt-based data generation and annotation pipeline, leveraging advanced LLMs as scalable generators, followed by rigorous verification by clinical experts. This process can be categorized into four primary paradigms.

\textbf{Paradigm 1: Structured Knowledge Probing.} This paradigm was employed to create datasets for assessing core medical knowledge (\textit{MAExam}, \textit{MAHalt}), safety awareness (\textit{MASafety}), and patient-facing service capabilities (\textit{MAKnowledge}, \textit{MAGuidance}, \textit{MAReport}, \textit{MAMedication}). The core methodology employed a suite of LLMs—namely Gemini-2.5-pro, Qwen3-235B-A22B, and DeepSeekR1-0528-671B—for the automated generation of medical examination items. For instance, to generate items for \textit{MAReport}, the LLM was provided with an authentic, anonymized clinical report and a structured prompt instructing it to formulate a clinically relevant multiple-choice question, generate one unambiguously correct answer, and create four plausible yet incorrect distractors that target common misconceptions.

\textbf{Paradigm 2: Scenario-Based Assessment of Safety and Ethics.} To evaluate model alignment with legal and ethical standards, we developed datasets comprising complex, open-ended scenarios (\textit{GeneralSafety}, \textit{MACompliance}, \textit{MAEthic}). The process began with the curation of challenging scenarios by medical and legal experts. These scenarios were framed as prompts to which a model generated a response. Subsequently, a separate set of structured, multi-dimensional evaluation prompts was used to score the generated response, assessing criteria such as compliance and risk awareness.

\textbf{Paradigm 3: High-Fidelity Information Extraction from Clinical Narratives.} To assess nuanced information extraction, we constructed four specialized datasets: \textit{MAEESymp} (symptoms), \textit{MAEEDisease} (diseases), \textit{MAEEMedic} (medications), and \textit{MAEEOper} (operations). The process started with a large corpus of anonymized Chinese electronic health records. We employed a semi-automated annotation pipeline where an LLM performed an initial pass of entity labeling, followed by a two-stage manual review and correction process by our physician team to ensure gold-standard accuracy.

\textbf{Paradigm 4: Simulation of Clinical Reasoning Tasks.} This category of datasets (\textit{MAInspection}, \textit{MADiagnosis}, \textit{MARecord}) was designed to simulate complex clinical reasoning, derived from authentic physician-patient dialogues and inpatient records sourced from web-based healthcare interactions. For instance, the construction of the \textit{MADiagnosis} dataset required clinical experts to sift through multiple, often conflicting, patient records to identify the most salient information for formulating a diagnosis, thereby simulating the critical real-world skill of evidence synthesis from complex and noisy data sources. This methodology ensures that the tasks and their corresponding ground truths are firmly rooted in actual clinical practice.

\subsubsection{Clinical Department Annotation}
To enable specialty-specific analysis, we implemented a systematic department annotation protocol for seven datasets within the \textit{SmartServ} and \textit{SmartCare} domains. We utilized a large language model to map each data instance to a standardized, two-tiered system of 9 primary and 64 secondary clinical departments. The annotation process adhered to strict rules: a maximum of three departments per instance, restriction to the predefined department list, and a ``No Department" label for non-clinical queries. This standardized annotation enhances the benchmark's utility for evaluating applications like automated patient department guidance.

\subsubsection{Expert Review and Refinement}
To ensure clinical fidelity, we instituted a multi-stage quality assurance protocol led by a panel of 300 licensed physicians from Grade-A tertiary hospitals in China, including experts with experience in web-based medical services. Our quality assurance involved a dual-annotator and cross-verification workflow. Two physicians independently reviewed each sample against five core criteria: natural and plausible, clinical validity, answer correctness, distractor plausibility (for multiple-choice questions), and rubric matches question. Items that failed this rigorous review were either discarded or revised and re-validated, ensuring every item in the \benchmarkname{} is accurate, relevant, and of high quality.

\subsection{Difficulty Classification}
To enable a fine-grained analysis of model capabilities, each question in the benchmark was assigned a difficulty level (Easy, Medium, Hard). This classification was determined empirically based on the performance of a panel of five diverse calibration LLMs (Mistral Large 2~\cite{jiang2023mistral7b}, Qwen2.5-72B~\cite{qwen2025qwen25technicalreport}, MedGemma-27B~\cite{sellergren2025medgemmatechnicalreport}, Baichuan-M1-14B~\cite{wang2025baichuanm1pushingmedicalcapability}, Citrus1.0-Qwen-72B~\cite{wang2025citrusleveragingexpertcognitive}). We employed three distinct methodologies tailored to different question types.
\begin{enumerate}
 \item \textbf{Objective Multiple-Choice Questions:} Difficulty was based on model consensus. A question was classified as Easy if all 5 models answered correctly, Medium if 4 models answered correctly, and Hard if 3 or fewer models answered correctly.
 \item \textbf{Open-Ended Questions with Standard Answers:} For tasks like Named Entity Recognition (NER), we used the micro-F1 score. A response was considered correct if its micro-F1 score exceeded a threshold of 0.8. Difficulty was then assigned using the same consensus logic as for multiple-choice questions.
 \item \textbf{Open-Ended Questions without Standard Answers:} For tasks in MedSE and SmartCare, we used a dual-model scoring system where two instances of our scoring model (GPT-4o-1120) independently scored each response on a scale of 0 to 1. If the scores diverged significantly (by more than 0.25), the response was flagged for final judgement by a human medical expert. The final average score across the five calibration models determined the difficulty: Easy ($>$0.7), Medium (0.6-0.7), and Hard ($<$0.6).
\end{enumerate}
This data-driven approach to difficulty scoring allows for a nuanced diagnosis of model weaknesses.

\subsection{Sample Selection}
To ensure \benchmarkname{} is comprehensive and robust, we designed a systematic sampling strategy guided by two principles: comprehensive coverage of clinical specialties/attributes and a controlled difficulty distribution. Table~\ref{tab:dataset_overview} details the sampling attributes and data counts.

For datasets with explicit specialty labels (e.g., \textit{MADiagnosis},\\ \textit{MARecord}), we used stratified sampling. We partitioned data into 64 clinical specialties and sampled proportionally from each stratum, ensuring broad coverage.
For datasets lacking specialty labels but having other key attributes (e.g., \textit{MASafety} by risk type, \textit{MAHalt} by hallucination type), we applied proportional sampling based on the original distribution of attributes to ensure diverse assessment dimensions.

To ensure discriminatory power, we implemented a multi-step iterative process to achieve a target difficulty distribution of 3:5:2 (Hard:Medium:Easy), chosen to effectively differentiate model capabilities. The process was as follows:
\begin{enumerate}
    \item \textbf{Initial Sampling and Profiling:} We sampled a preliminary set based on coverage principles and analyzed its difficulty profile.
    \item \textbf{Gap Analysis and Supplementation:} We calculated the shortfall against our 3:5:2 target and performed targeted sampling from the remaining data pool to fill these gaps, prioritizing under-represented categories.
    \item \textbf{Balancing and Verification:} We made fine-tuning adjustments (e.g., swapping items of the same specialty and difficulty) to precisely match the target ratio. The final set was verified to adhere to both coverage and difficulty principles.
\end{enumerate}

\subsection{Evaluation Methodology} \label{sec:eval_methodology}
Our evaluation protocol was designed to be rigorous, fair, and efficient, incorporating a systematic sampling strategy, a mechanism to mitigate bias, and a hybrid scoring framework centered on our novel SFT-based judge model.

\subsubsection{Dynamic Evaluation for Bias Mitigation}
To mitigate positional bias in multiple-choice questions, we implemented a dynamic shuffling strategy. Each question was presented to the model three times, with the order of options randomized in each trial. A question was marked as correct only if the model provided the semantically correct answer across all three permutations. This method ensures that models reason about the content of the options rather than relying on positional heuristics.

\subsubsection{Hybrid Scoring Protocol}
We developed a hybrid scoring protocol that combines automated metrics for structured tasks with an AI-assisted, expert-adjudicated framework for complex, open-ended questions.
\begin{table}[h]
 \caption{List of LLMs Evaluated in This Study. The table lists the 10 diverse Large Language Models evaluated, including their originating organization, access type (Open-source/Closed-source), and reported parameter size (where available).}
 \label{tab:models_evaluated}
 \begin{tabular}{l l l r}
 \toprule
 \textbf{Model} & \textbf{Organization} & \textbf{Type} & \textbf{Parameters} \\
 \midrule
 Kimi-K2-Instruct \cite{kimiteam2025kimik2openagentic} & Moonshot AI & Closed & $\sim$1000B \\
 DeepSeekR1\_0528 \cite{deepseekai2025deepseekr1incentivizingreasoningcapability} & DeepSeek & Open & 671B \\
 Claude-4-Sonnet \cite{Anthropic_Claude4_2025} & Anthropic & Closed & N/A \\
 GPT-5 \cite{OpenAI2025GPT5} & OpenAI & Closed & N/A \\
 Baichuan-M2 \cite{m2team2025baichuanm2scalingmedicalcapability} & Baichuan & Open & 32B \\
 DeepSeek-V3 \cite{deepseekai2025deepseekv3technicalreport} & DeepSeek & Open & 671B \\
 Qwen3-235B \cite{yang2025qwen3technicalreport} & Alibaba Cloud & Open & 235B \\
 GPT-4o \cite{openai2024gpt4technicalreport} & OpenAI & Closed & N/A \\
 Citrus1.0 \cite{wang2025citrusleveragingexpertcognitive} & JD.com & Open & 72B \\
 Llama 3.3 \cite{grattafiori2024llama3herdmodels} & Meta AI & Open & 70B \\
 \bottomrule
 \end{tabular}
\end{table}

\paragraph{Fully Automated Evaluation}
For tasks with objective, verifiable answers, scoring was fully automated.
\begin{itemize}
 \item \textbf{Accuracy for Multiple-Choice Questions:} This metric was used for all multiple-choice datasets. A response was scored as correct only if it matched the ground truth across all three shuffled trials.
 \item \textbf{Micro-F1 Score for Structured Extraction:} This metric was applied to tasks like Named Entity Recognition. The Micro-F1 score was calculated by aggregating true positives, false negatives, and false positives across all entity classes.
\end{itemize}
\paragraph{AI-Assisted Evaluation with Expert judgement}
For open-ended questions lacking a single ground truth, we employed a dual-assessor AI system with a human-in-the-loop workflow.
\begin{itemize}
 \item \textbf{Initial AI Assessment:} Each response was independently evaluated by two separate instances of our scoring model (GPT-4o-1120) against a predefined, multi-dimensional scoring rubric.
 \item \textbf{Human Judgement Trigger:} A manual review by a human medical expert was automatically triggered if the scores from the two AI assessors exhibited a discrepancy greater than 0.25 on a normalized 0-1 scale.
 \item \textbf{Expert Review Workflow:} When triggered, such cases were initially escalated to a physician from our expert panel for definitive assessment. 
\end{itemize}

While this human-in-the-loop framework ensures gold-standard accuracy, it introduces a significant evaluation bottleneck. To address this challenge of scalability and cost, we developed our automated judge model as a third-stage assessor.

\paragraph{Automated Judge Model}
A fundamental barrier to evaluating complex, open-ended scenarios at scale is the reliance on costly and subjective human expert scoring. To overcome this, we introduce a specialized ``judge" model as a core component of our evaluation methodology. This model is engineered to serve as a consistent, scalable, and expert-aligned adjudicator for disputed or ambiguous responses. We trained this model using Supervised Fine-Tuning (SFT) on a substantial corpus of 19,000 expert-annotated scoring points, derived from complex cases where high-capability generalist models disagreed. This SFT process imparts the model with the nuanced decision-making capability of clinical experts, enabling a reproducible and cost-effective evaluation pipeline.
 The training corpus was constructed by aggregating the expert-annotated scoring points from the disputed questions generated during the evaluation of nine distinct LLMs. This dataset captures a wide array of complex and ambiguous cases.
The evaluation set was derived from the complete set of 596 disputed questions generated by multiple models that was held out from the training data, ensuring a rigorous test of generalization. This specialized SFT model is already in preliminary online deployment, providing decision-support assistance to clinicians and validating its real-world utility.

\vspace{-4mm}
\section{Experimental Results}
\subsection{Evaluated Models}
We evaluated a diverse set of prominent LLMs on the \benchmarkname{} to assess the current state-of-the-art. The selection included both open-source and closed-source models of varying sizes, representing a cross-section of the field's leading contenders. The models evaluated are listed in Table~\ref{tab:models_evaluated}.                 
All models were evaluated using their official API endpoints. To ensure the integrity of the evaluation and prevent any potential data contamination, the models were not provided with any information about the questions, tasks, or answer formats beforehand, and all interactions were conducted via their respective APIs.

\vspace{-2mm}

\begin{table*}[t]
  \centering
  \caption{Overall Performance of Evaluated LLMs Across the Five Core Dimensions of the \benchmarkname{}. Scores represent the mean performance. The best-performing model in each column is in \textbf{bold}, and the second-best is \underline{underlined}.}
  \label{tab:llm_performance}
  \begin{tabular}{l c c c c c c}
    \toprule
    \textbf{Model} & \textbf{Average} & \textbf{MedKQA} & \textbf{MedSE} & \textbf{MedRU} & \textbf{SmartServ} & \textbf{SmartCare} \\
    \midrule
    Kimi-K2-Instruct~\cite{kimiteam2025kimik2openagentic} & \textbf{77.288} & 73.17 & \underline{85.50} & 87.75 & \textbf{61.25} & \underline{78.77} \\
    DeepSeekR1-0528-671B~\cite{guo2025deepseek} & \underline{75.434} & \textbf{78.58} & 85.09 & \underline{88.27} & 53.33 & 71.90 \\
    Claude-4-Sonnet~\cite{Anthropic_Claude4_2025} & 74.534 & \underline{76.42} & 73.67 & \textbf{88.78} & \underline{58.42} & 75.38 \\
    GPT-5~\cite{OpenAI2025GPT5} & 73.87 & 76.00 & 77.80 & 88.02 & 55.42 & 72.11 \\
    Baichuan-M2-32B~\cite{m2team2025baichuanm2scalingmedicalcapability} & 72.012 & 65.25 & \textbf{90.64} & 83.60 & 49.67 & 70.90 \\
    DeepSeek-V3~\cite{deepseekai2025deepseekv3technicalreport} & 70.498 & 62.08 & 83.92 & 87.04 & 49.75 & 69.70 \\
    Qwen3-235B-A22B~\cite{yang2025qwen3technicalreport} & 65.886 & 52.00 & 80.61 & 85.25 & 32.58 & \textbf{78.99} \\
    GPT-4o~\cite{openai2024gpt4technicalreport} & 65.45 & 49.25 & 73.57 & 85.86 & 44.83 & 73.74 \\
    Citrus1.0-Qwen-72B~\cite{wang2025citrusleveragingexpertcognitive} & 63.946 & 67.50 & 70.95 & 78.14 & 33.67 & 69.47 \\
    Llama 3.3-70B~\cite{grattafiori2024llama3herdmodels} & 54.022 & 42.75 & 58.70 & 79.74 & 32.67 & 56.25 \\
    \bottomrule
  \end{tabular}
\end{table*}

\subsection{Overall Performance Analysis}
As shown in Table~\ref{tab:llm_performance}, a clear performance hierarchy emerged. Kimi-K2-Instruct achieved the highest overall score of 77.29, demonstrating robust and well-rounded capabilities. It was closely followed by DeepSeekR1-0528-671B (75.43) and Claude-4-Sonnet (74.53). A critical observation is the universal performance drop in the SmartServ (Smart  Services) dimension. Although Kimi-K2-Instruct leads this category, its score of 61.25 is significantly lower than scores in other dimensions, and most models score below 55. This indicates a substantial gap between foundational knowledge and the ability to apply it in interactive, patient-facing scenarios. In contrast, performance on MedRU (Medical Record Understanding) was consistently high across top models, suggesting that fundamental NLP tasks like information extraction are relatively mature. Performance on MedSE (Medical Safety \& Ethics) was also strong, with Baichuan-M2-32B achieving an exceptional 90.64, indicating highly effective alignment training.

\subsection{Performance on Sub-dimension SmartServ}
\begin{figure}[h]
 \centering
 \includegraphics[width=\linewidth]{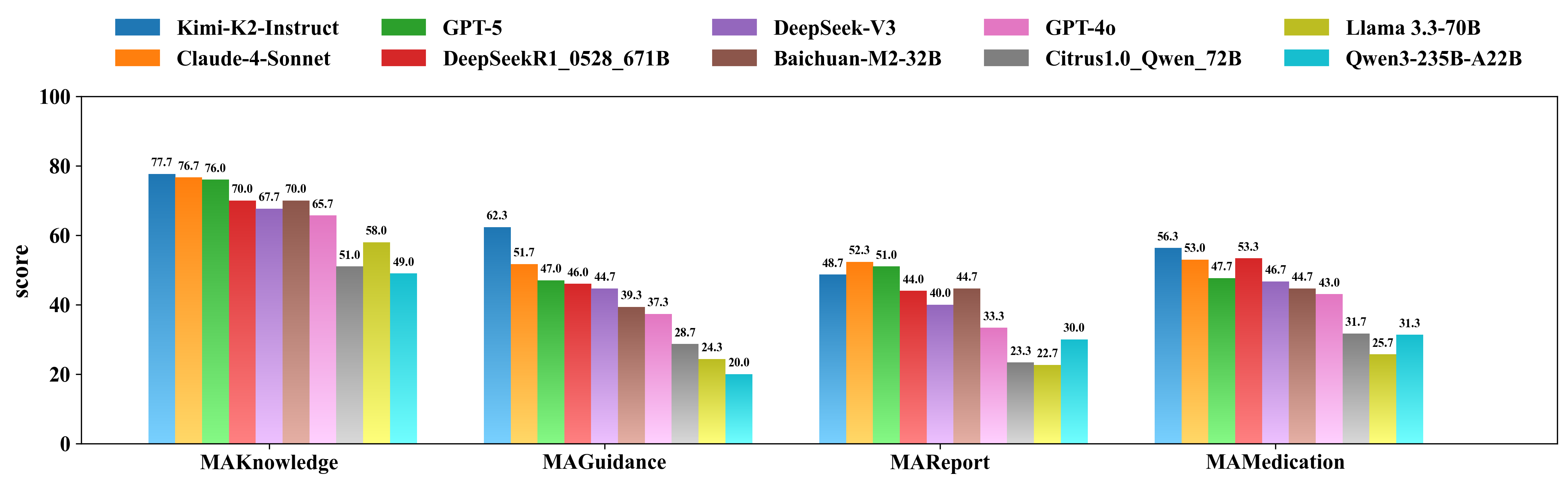}
 \caption{Bar chart comparing the performance of selected models on the SmartServ sub-dimension of the \benchmarkname{}.}
 \label{fig:radar_chart}
\end{figure}

To provide a more granular view, Figure~\ref{fig:radar_chart} visualizes the performance of top models on  SmartServ. This reveals specific strengths and weaknesses masked by aggregated scores. The results highlight that even top-performing models have uneven capability profiles. For example, Kimi-K2-Instruct showed superior performance, ranking first in the MAKnowledge dimension.  Notably, Baichuan-M2, despite its smaller size, ranked fourth in the MAKnowledge. 

\vspace{-6mm}
\subsection{Judge Model Performance}
Validating our SFT-based judge model is a prerequisite for our main findings (Section 5.1). We confirmed it can reliably resolve scoring discrepancies, surpassing general-purpose LLMs. The results (Table~\ref{tab:adjudication_results}) demonstrate three key findings.

\begin{table}[h]
  \caption{Performance of the Automated Judge Model on Disputed Cases, validating the SFT approach.}
  \label{tab:adjudication_results}
  \resizebox{\linewidth}{!}{%
    \begin{tabular}{l c c c c c}
    \toprule
    \textbf{Model} & \textbf{Accuracy} & \textbf{Precision} & \textbf{Recall} & \textbf{F1-Score} & \textbf{Cohen's Kappa} \\
    \midrule
    Qwen3-235B-A22B, w/o SFT & 0.7772&0.9416&    0.7160&    0.8135&    0.5493 \\
    DeepSeek-V3, w/o SFT & 0.8677&0.8559&    0.9679&    0.9085&    0.6730 \\
    \midrule
    Kimi-K2-Instruct-0905, w/o SFT&    0.8007&    0.8095&    0.9235&    0.8627&    0.5045\\
    Gemini-2.5-pro, w/o SFT&    0.7588&    0.8872&    0.7383&    0.8059&    0.4943\\
    GLM-4.6, w/o SFT    &0.7513&    0.9103&    0.7030&    0.7933&    0.4937\\
    GPT-5-2025-08-07, w/o SFT&    0.6633    &0.9435    &0.5358    &0.6835    &0.3776\\
    \midrule
    Qwen3-14B, w/o SFT & 0.7102    &0.8946&    0.6494&    0.7525    &0.4235\\
    Qwen3-14B, with SFT on 2k data & 0.8325    &0.8458    &0.9210&    0.8818    &0.5962 \\
    \textbf{Qwen3-14B, with SFT 19k data} &\textbf{0.9213}    &\textbf{0.9163}&    \textbf{0.9728}    &\textbf{0.9437}    &\textbf{0.8132} \\
    \bottomrule
    \end{tabular}
  } 
\end{table}

First, our SFT-trained Qwen3-14B (trained on 19k data) achieves state-of-the-art 92.13\% accuracy and 94.37\% F1-Score, significantly outperforming all general-purpose baselines, including much larger models like DeepSeek-V3 and GPT-5. Notably, the model maintains consistent performance despite uneven input lengths (ranging from 764 to 1728 tokens). We utilized Qwen3-14B as the backbone because its compact parameter size, combined with robust performance, makes it a highly viable solution for practical clinical deployment.

Second, this performance is a direct result of specialized SFT. An ablation study confirms this: the base Qwen3-14B accuracy (71.02\%) improved to 83.25\% (with 2k SFT samples) and finally to 92.13\% (with 19k samples), validating that the model learns specialized capabilities from the expert-annotated data.

Third, the model achieved a Cohen's Kappa of \textbf{81.32\%}, signifying high agreement with human experts. This confirms the model's judgments are trustworthy and aligned with human reasoning, unlike the general-purpose models. Our SFT-based methodology thus provides a robust and trustworthy proxy for human-level evaluation, enabling the scalable analysis in this paper.

\section{Discussion}
The results from our evaluation on the \benchmarkname{} provide several critical insights into the landscape of medical LLMs.

\textbf{Model Scale and Specialization Drive Performance:} The top-ranking performance of Kimi-K2-Instruct, a model with a very large parameter count, suggests scale remains a significant factor in achieving high overall competence. Concurrently, the exceptional safety score of the much smaller Baichuan-M2-32B highlights that targeted training and fine-tuning are equally critical. The relatively modest scores of powerful generalist models like GPT-4o and Llama 3.3 further underscore that general-purpose excellence does not guarantee proficiency in the high-stakes medical domain.

\textbf{Clear Evidence of Progress Within Model Families:} Our results demonstrate significant iterative improvement within model series. For example, GPT-5 (73.87) substantially outperforms GPT-4o (65.45) across nearly all dimensions, indicating effective advancements in architecture and training. A similar leap is observed from DeepSeek-V3 (70.50) to DeepSeekR1-0528-671B (75.43), showcasing the rapid pace of progress in the field.

\textbf{Scenario Application Bottleneck:} The most striking finding is the universal difficulty models face in the SmartServ dimension. This category requires more than simple knowledge retrieval; it demands contextual understanding and the ability to interpret user intent from non-technical language. Tasks like report interpretation and Department Guidance are complex reasoning processes that current LLMs struggle to execute reliably. This translational gap is the most significant challenge for the practical deployment of medical LLMs in patient-facing roles.

\textbf{Unpacking Diverse Architectural Strengths:} The \benchmarkname{} allows for a nuanced analysis of each model's unique profile.
\begin{itemize}
 \item \textbf{Kimi-K2-Instruct} is the best all-rounder, with a leading overall score for the challenging patient-facing (SmartServ) tasks.
 \item \textbf{DeepSeekR1-0528-671B} excels in foundational knowledge, achieving the top score in MedKQA, making it a strong candidate for knowledge-intensive applications.
 \item \textbf{Claude-4-Sonnet} shows exceptional strength in structured data processing, leading the MedRU dimension, which is critical for tasks involving electronic health records.
 \item \textbf{Baichuan-M2-32B}, despite its smaller size, presents a superior performance in safety alignment with its category-defining score of 90.64 in MedSE.
 \item \textbf{Qwen3-235B-A22B} demonstrates a surprising specialization, securing the top position in SmartCare, suggesting it is highly optimized for physician-support tasks.
\end{itemize}

\vspace{-1mm}
\textbf{A Methodological Contribution via SFT:} Beyond the model performance insights, our work presents a significant methodological contribution: the validation of an SFT-trained judge model for scalable, high-fidelity benchmark evaluation. The failure of even powerful generalist models (like DeepSeek-V3 and GPT-5) to match our specialized judge's accuracy (92.13\%) on disputed cases (Table~\ref{tab:adjudication_results}) proves that ``judging" nuanced medical responses is a distinct, complex skill. Our SFT approach provides a blueprint for future benchmarks to move beyond costly human-in-the-loop evaluation, enabling the field to reliably assess the scenario-based capabilities.

\textbf{Implications and Future Directions:} Our findings suggest that the future of medical LLM development must prioritize scenario-based training. This involves moving beyond static Q\&A datasets to incorporate more interactive, dialogue-based tasks that mimic real clinical workflows. Furthermore, the development of specialized evaluation metrics for patient-facing communication is crucial. The \benchmarkname{} will continue to evolve, incorporating more complex clinical scenarios to push the frontier of medical AI evaluation.


\vspace{-3mm}
\section{Conclusion}
In this work, we introduced \benchmarkname{}, a comprehensive benchmark designed to rigorously evaluate LLMs in clinical contexts. Through our multi-dimensional assessment of leading models, we revealed critical gaps in safety and reliability that traditional metrics overlook. By leveraging our granular, scenario-driven evaluation benchmark to expose LLMs' issues in the perspective of safety, ethics, and compliance, this work bridges the trust gap between technical capability and clinical reality, ensuring that future AI agents evolve from experimental tools into trustworthy partners that safely empower physicians and democratize high-quality patient care, adhering to the vision of AI for social good.


\vspace{-3mm}
\section{Ethical Considerations}
The research protocol was approved by the Institutional Review Board (IRB) under the protocol number 2025-074. This study was conducted in strict adherence to the ACM Publications Policy on Research Involving Human Participants and Subjects. To ensure privacy and confidentiality, all data were fully de-identified prior to their use in this research.

\bibliographystyle{ACM-Reference-Format}
\bibliography{main}
\appendix
\end{document}